\pdfoutput=1

\documentclass[11pt]{article}

\usepackage{emnlp2021}

\usepackage{times}
\usepackage{latexsym}

\usepackage{graphicx} 
\usepackage{times}
\usepackage{latexsym}
\usepackage{pxrubrica}        
\usepackage{tabularx}
\usepackage{url}
\usepackage{booktabs}
\usepackage{multirow}
\usepackage{arydshln}
\usepackage{amsmath}
\usepackage{xcolor}

\usepackage[T1]{fontenc}

\usepackage[utf8]{inputenc}

\usepackage{microtype}

\title{Mitigation of Diachronic Bias in Fake News Detection Dataset}

\author{Taichi Murayama \and Shoko Wakamiya \and Eiji Aramaki \\
Nara Institute of Science and Technology (NAIST) \\
\texttt{\{murayama.taichi.mk1, wakamiya, aramaki\}@is.naist.jp}
}

\begin{document}
\maketitle
\begin{abstract}
Fake news causes significant damage to society.
To deal with these fake news, several studies on building detection models and arranging datasets have been conducted.
Most of the fake news datasets depend on a specific time period.
Consequently, the detection models trained on such a dataset have difficulty detecting novel fake news generated by political changes and social changes;
they may possibly result in biased output from the input, including specific person names and organizational names.
We refer to this problem as \textbf{Diachronic Bias} because it is caused by the creation date of news in each dataset.
In this study, we confirm the bias, especially proper nouns including person names, from the deviation of phrase appearances in each dataset.
Based on these findings, we propose masking methods using Wikidata to mitigate the influence of person names and validate whether they make fake news detection models robust through experiments with in-domain and out-of-domain data.
\end{abstract}
\section{Introduction}
Fake news, which refers to intentional and verifiable false news stories, has caused significant damage to society.
For example, ~\citet{2016election2} noted that 25\% of the news stories linked in tweets posted just before the 2016 U.S. presidential election were either fake or extraordinarily biased.
In addition to elections, fake news tends to spread after unusual situations such as natural disasters~\cite{hashimoto2020} and disease outbreaks (e.g., COVID-19~\cite{fakecovid}).
To address these problems, various studies on the development of models for fake news detection from social media posts and news content, and the construction of fake news datasets for this purpose have been conducted~\cite{survey1}.

Most datasets for fake news detection consist of factual and fake news that actually diffuse over the Internet.
The topics and contents of fake news change over time because they are strongly influenced by the interests of the general population~\cite{pnas2017}.
For instance, there were fake news related to President Obama in 2013~\cite{obama}, presidential election in 2016~\cite{2016election2}, and COVID-19 in 2020~\cite{covid-2}.
Thus, datasets are frequently constructed based on fake news present in a specific period.
Fake news detection models learned from these datasets achieve high accuracy for the datasets constructed for the same period and domain, while they lead to a drop in the detection performance of fake news in different domains and future applications because of the difference in word appearance. 
In other words, fake news detection models learned from a dataset that includes news only from a specific period may possibly result in a biased judgment from the input, based on cues related to a particular person name or an organizational name.
For example, a model learned from a dataset in 2017 is difficult to correctly classify articles including ``Donald Trump'' or ``Joe Biden'' in 2021, because the model does not know a new president change.
We call the problem as \textbf{Diachronic Bias} because it is caused by the difference in the news publishing date in each dataset.
This problem occurs with data from even the same domain, making it difficult to construct a robust detection model.

\begin{table*}[!]
    \centering
    \scriptsize
    \begin{tabular}{lrrlrr lrrlrr} \toprule
         \multicolumn{6}{c}{MultiFC} & \multicolumn{6}{c}{Constraint}\\ \cmidrule(lr){1-6} \cmidrule(lr){7-12}
         \multicolumn{3}{c}{Real} & \multicolumn{3}{c}{Fake} &
         \multicolumn{3}{c}{Real} & \multicolumn{3}{c}{Fake}\\ \cmidrule(lr){1-3} \cmidrule(lr){4-6} \cmidrule(lr){7-9} \cmidrule(lr){10-12}
         \textbf{Bi-gram} & \textbf{LMI} & $p(l|w)$ & \textbf{Bi-gram} & \textbf{LMI} & $p(l|w)$ & \textbf{Bi-gram} & \textbf{LMI} & $p(l|w)$  & \textbf{Bi-gram} & \textbf{LMI} & $p(l|w)$\\ \cmidrule(lr){1-3} \cmidrule(lr){4-6} \cmidrule(lr){7-9} \cmidrule(lr){10-12}
         \textbf{mitt romney} & 218 & 0.69 & health care & 631 & 0.64 & url url & 1378 & 0.77 & a video & 591 & 1.0\\
         if you & 217 & 0.70 & \textbf{barack obama} & 365 & 0.69 & rt @user & 822 & 0.93 & \textbf{donald trump} & 569 & 0.98\\
         rhode island & 190 & 0.75 & \textbf{president barack} & 337 & 0.70 & total number & 650 & 0.98 & has been & 569 & 0.52\\
         new jersey & 177 & 0.67 & \textbf{scott walker} & 258 & 0.81 & more than & 635 & 0.89 & \textbf{url donaldtrump} & 435 & 1.0\\
         \textbf{john mccain} & 167 & 0.73 & says president & 218 & 0.78 & have been & 575 & 0.82 & \textbf{bill gates} & 355 & 1.0\\
         no. 1 & 128 & 0.86 & care law & 185 & 0.80 & @user url & 449 & 0.87 & video shows & 346 & 0.98\\
         voted against & 128 & 0.71 & will be & 162 & 0.63 & managed isolation & 402 & 1.0 & \textbf{president trump} & 315 & 1.0\\
         any other & 125 & 0.61 & \textbf{hillary clinton} & 159 & 0.67 & our daily & 385 & 0.99 & covid vaccine & 293 & 0.80\\
         does not & 119 & 0.71 & \textbf{gov. scott} & 148 & 0.72 & states reported & 373 & 1.0 & corona virus & 275 & 1.0\\
         this year & 116 & 0.75 & social security & 144 & 0.68 & update published & 367 & 1.0 & social media & 275 & 0.93\\
         \bottomrule
    \end{tabular}
    \caption{Top 10 LMI ranked bi-grams in MultiFC and Constraint for real and fake labels with their $p(l | w)$.
    \textbf{LMI} are written as value multiplied by $10^{6}$.
    Person names are written in bold. 
    There is a tendency for real labels to be highly correlated with common phrases, while fake labels are highly correlated with person names.}
    \label{LMI}
\end{table*}

This study examines various strategies to mitigate diachronic bias by masking proper nouns that tend to cause the bias, such as names of people and places.
First, we analyzed the correlation between labels and phrases in several fake news detection datasets with different creation periods and noted the deviation of words, mainly person names.
We then applied and validated several masking methods focusing on proper nouns to mitigate diachronic bias in the tackling of fake news detection tasks.

\section{Related Work}
Analysis and examination of mitigation methods for various types of bias have been conducted for detecting offensive language and hate speech: author bias~\cite{author_bias}, annotator bias~\cite{annotate_bias}, gender bias~\cite{gender_bias1,gender_bias2}, racial bias~\cite{racial_baias1}, political bias~\cite{poliotics}, etc.
\citet{masking_actor} focuses on frequency bias of person name in their dataset, while we handle person name considering the passage of time.

In various studies, bias analysis and mitigation are addressed for the fact verification task, where given texts as judged as factual or otherwise from several pieces of evidence, and is one of the recognizing textual entailment tasks.
For example, ~\citet{schuster} and  ~\citet{suntwal2019importance} proposed mitigation methods by replacing some words with specific labels to build a robust inference model for out-of-domain data.
However, to the best of our knowledge, there has been no study or analysis of bias in fake news detection datasets.

\section{Resources}

\subsection{Datasets}
We examine diachronic bias by analyzing four fake news detection datasets with different domains and creation periods.
Each article and post in these datasets has a binary label (real/fake).
The details of the datasets are as follows and further information is provided in Appendix~\ref{dataset}:

\textbf{MultiFC}~\cite{augenstein2019multifc}: This is a multi-domain dataset containing over 36,000 headlines from 38 fact-checking organizations.
We extracted 7,861 headlines from 2007 to 2015 and regarded headlines labeled as ``truth!,'' ``true,'' or ``mostly true'' as real; those labeled ``mostly false'' or ``false'' as fake.

\textbf{Horne17}~\cite{horne2017just}: This dataset contains news articles on the 2016 US presidential election, which are labeled real/fake/satire, based on the investigation of BuzzFeed News. 
We used articles with fake and real labels.

\textbf{Celebrity}~\cite{automatic}: This dataset is composed of news articles verified by Gossipcop, which targets news related to celebrities.
Most of the articles, whose topics are mainly sensational, such as fights between celebrities, were published between 2016 and 2017.

\textbf{Constraint}~\cite{patwa2020fighting}: This dataset was used in the CONSTRAINT 2021 shared task and consists of social media posts related to COVID-19.
These posts were verified by fact-checking sites such as Politifact and Snopes.

\subsection{Correlation between phrases and labels}
We investigated the correlation between phrases and labels to examine bias in each dataset.
To capture high-frequency phrases that are highly correlated with a particular label, we use local mutual information (LMI)~\cite{evert2005statistics}.
Given a dataset $D$, the LMI between a phrase $w$ and label $l$ is defined as follows:
$
LMI(w, l)  = p(w,l) \cdot \log \bigg(\frac{p(l | w)}{p(l)}\bigg),
$
where $p(w,l)$ is calculated as $\frac{\operatorname{count}(w,l)}{|P|}$, $p(l | w)$ as $\frac{\operatorname{count}(w, l)}{\operatorname{count}(w)}$, $p(l)$ as $\frac{\operatorname{count}(l)}{|P|}$, and $|P|$ is the number of occurrences of all phrases in $D$.

Table~\ref{LMI} presents bi-grams that are highly correlated with each label in MultiFC and Constraint\footnote{Appendix~\ref{lmi} lists bi-grams with high LMI in Horne17 and Celebrity.}.
In MultiFC, the headlines prior to 2015 containing words referring to the U.S. president at the time, such as ``barack obama,'' were highly correlated with fake labels.
In Constraint, words such as ``bill gates'' and ``donald trump,'' were highly correlated with the fake labels.
Some person names were highly correlated with real labels, but these LMI values were much lower than those that were highly correlated with the fake labels.
These results revealed a bias in the relationship between specific person names and labels.
The detection model trained on one of these datasets is not adaptable to instances such as the change of president, and thus does not work well on other datasets.


\section{Diachronic bias mitigation}

\subsection{Masking methods}
We examine multiple masking methods, starting from word deletion to word replacement for input text data to mitigate the diachronic bias and to build a robust detection model for out-of-domain data.
We utilize Named Entity Recognition (NER)~\cite{akbik-etal-2019-pooled} in Flair~\cite{flair} to search for words to be used as masks.
Examples of each masking method are listed in Appendix~\ref{masking_ref}.

\textbf{Named Entity (NE) Del}: 
Words tagged with NEs are removed.
This masking method aims to build a detection model independent of NE.

\textbf{Basic NER}: 
Words tagged with NEs are replaced with the corresponding labels, such as PER (person label), LOC (location label), etc.

\textbf{WikiD}: 
Words tagged with PER labels are replaced with Wikidata~\cite{Wikidata} label, specifically, position held (P39), or alternatively occupation (P106) depending on availability.
For example, the use of Wikidata at that time made it possible to replace the phrase ``barack obama’’ in articles from 2015 and ``donald trump’’ in those from 2020 with the same label as President of the United States (Q11696).
This makes fake news detection models more robust against the passage of time and potentially more effective in mitigating the bias.

\textbf{WikiD+Del}: 
Words tagged with PER labels are replaced by the same rule as WikiD, and words tagged with other NEs are removed.

\textbf{WikiD+NER}: 
Words tagged with PER labels are replaced by the same rule as WikiD, and words tagged with other NEs are replaced with the corresponding label.

\subsection{Experimental Setting}
We verify the effectiveness of masking methods for fake news detection.
We examine how well the detection models perform with each of the masking methods against in-domain and out-of-domain for all datasets.
Our in-domain experiments mainly investigate the effect of each masking method on accuracy in the same domain.
Our out-of-domain experiments validate the effect of each masking method in datasets with considering the flow of time.
We consider that out-of-domain setting is close to reality and useful whether each masking method is effective against diachronic bias.

\textbf{Model}: Our experiments utilize a pretrained model, $\mathrm{BERT}_{\mathrm{BASE}}$ model~\cite{DBLP:journals/corr/abs-1810-04805}, which is made freely available by Google.
Labels (LOC, Q11696, etc.) replaced by each masking method were handled as new tokens during the fine-tuning of the pretrained model.

\textbf{Data and Evaluation}: 
Each dataset is randomly divided into training (80\%) and test (20\%) sets.
The time-based splitting is suitable for our experimental settings than the random splitting in each dataset.
However, it was difficult for us to apply the time-based splitting for in-domain experiment because the published time is not described in most of the samples\footnote{In appendix \ref{time_ref}, we also conduct an experiment using MultiFC, which contains the published time, with a time-based splitting in train and test sets for trying to remove the effect of domain shift.}.
Out-of-domain experimental settings mean the same verification as the time-based splitting.
For example, the evaluation of models, trained in MultiFC consisting of events in 2015, on Constraint consisting of events in 2020 is equivalent to time-based splitting.

\subsection{Experimental Results and Discussions}

\subsubsection{In-domain data}
Table~\ref{in_domain} presents the accuracies of each masking method against the in-domain data.
No Mask, with no application of the masking method, achieved the highest accuracy in all datasets except Constraint.
On the other hand, WikiD achieved the highest accuracy on Constraint.
In addition, there was only a slight difference in accuracy between No Mask and the other masking methods, even in the datasets where No Mask had the highest accuracy.
These results suggest that these masking methods result in insignificant decrease in accuracy when tested with the in-domain dataset.

\begin{table}[t]
    \centering
    \footnotesize
    \scalebox{0.9}{
    \begin{tabularx}{8.6cm}{lrrrr} \toprule
    \multicolumn{1}{c}{\multirow{2}*{Method}} & \multicolumn{4}{c}{Test set}\\ \cmidrule{2-5}
     & MultiFC & Horne17 & Celebrity & Constraint\\ \midrule
    No Mask & \textbf{0.681} & \textbf{0.746} & \textbf{0.760} & 0.960\\
    NE Del & 0.656 & 0.706 & 0.750 & 0.959\\
    Basic NER & 0.659 & 0.735 & 0.750 & 0.950\\
    WikiD & 0.675 & 0.725 & 0.730 & \textbf{0.967}\\
    WikiD+Del & 0.660 & 0.706 & 0.700 & 0.959\\
    WikiD+NER & 0.660 & 0.640 & 0.730 & 0.957\\
    \bottomrule
    \end{tabularx}
    }
    \caption{\label{in_domain}
    Accuracy of each masking method against in-domain data. Bold indicate the highest accuracies.}
\end{table}

\subsubsection{Out-of-domain data}
Table~\ref{out_domain} presents the accuracies of each masking method against the out-of-domain data.
For almost all of the out-of-domain test data, each masking method achieved higher accuracy than No Mask.

Results of No Mask indicate the difficulty of adapting to out-of-domain data.
For example, the model trained on Constraint achieved a high accuracy of $0.967$ on Constraint (refer to Table~\ref{in_domain}); however, the model trained on datasets except Constraint achieved low accuracies, ranging from $0.48$ to $0.58$ on the same test set. 
These results imply the difficulty of generalizing the fake news detection model.
NE Del achieved the same of higher accuracy than No Mask in 9 out of 12 settings, although it is the simplest masking method.
In particular, NE Del trained on Horne17 achieved the highest accuracy for MultiFC and Celebrity among models trained on Horne17.
Although Basic NER trained on Horne17 also achieved the highest accuracy for Constraint, the improvement was smaller compared to that of NE Del.

Wikidata-based masking methods WikiD and WikiD+Del achieved higher accuracy compared to No Mask in 9 settings, except when testing with Celebrity test set.
In particular, the accuracies in WikiD have statistically significant improvements in 6 settings, compared to No Mask.
For example, the model trained on MultiFC achieved a significant improvement in accuracy from $0.530$ (No Mask) to $0.689$ (WikiD), against Constraint test set.
These results reveal that the masking method using Wikidata could mitigate the diachronic bias and build robust models even for out-of-domain data.
In appendix \ref{example_tweet}, we show some examples, which WikiD accurately classifies while No Mask makes the wrong classification.
However, against Celebrity test set, whose domain is entertainment, the accuracy of No Mask is almost the same as that of these masking methods.
We consider that this is due to the non-applicability of Wikidata owing to the difference in domain between Celebrity and other datasets; 
53.6\% of Wikidata labels of Horne17 were found in MultiFC, while only 33.6\% of those were found in Celebrity (refer to Appendix~\ref{wiki_ref}).
Additionally, WikiD+NER has comparatively lower accuracies than WikiD+Del.
This indicates that we can build a more robust model by removing entities other than person names.
We believe that the model can focus on stylistic features by removing extra entity information.

\begin{table}[t]
    \centering
    \footnotesize
    \scalebox{0.8}{
    \begin{tabularx}{10cm}{clrrrr} \toprule
    Train & \multicolumn{1}{c}{\multirow{2}*{Method}} & \multicolumn{4}{c}{Test set}\\ \cmidrule{3-6}
    set & & MultiFC & Horne17 & Celebrity & Constraint\\ \midrule
    \multicolumn{1}{c}{\multirow{6}*{\rotatebox[origin=c]{90}{MultiFC}}} &No Mask & - & 0.706 & \textbf{0.660} & 0.530\\
    &NE Del & - & 0.706 & 0.590 & *0.664\\
    &Basic NER & - & 0.725 & 0.600 & *0.680\\
    &WikiD & - & \textbf{0.746} & 0.590 & *\textbf{0.689}\\
    &WikiD+Del & - & 0.725 & \textbf{0.660} & *0.669\\
    &WikiD+NER & - & 0.632 & 0.520 & *0.667\\ \midrule
    \multicolumn{1}{c}{\multirow{6}*{\rotatebox[origin=c]{90}{Horne17}}} & No Mask & 0.504 & - & 0.670 & 0.481\\
    &NE Del & *\textbf{0.551} & - & \textbf{0.680} & *0.553\\
    &Basic NER & 0.523 & - & 0.670 & *\textbf{0.563}\\
    &WikiD & *0.525 & - & 0.620 & 0.487\\
    &WikiD+Del & 0.523 & - & 0.610 & 0.515\\
    &WikiD+NER & 0.500 & - & 0.630 & *0.531\\ \midrule
    \multicolumn{1}{c}{\multirow{6}*{\rotatebox[origin=c]{90}{Celebrity}}} & No Mask & 0.533 & 0.451 & - & 0.583\\
    &NE Del & 0.545 & 0.529 & - & *\textbf{0.763}\\
    &Basic NER & 0.521 & 0.549 & - & 0.568\\
    &WikiD & *\textbf{0.555} & *0.549 & - & *0.724\\
    &WikiD+Del & 0.534 & 0.529 & - & *0.663\\
    &WikiD+NER & 0.525 & *\textbf{0.568} & - & 0.598\\ \midrule
    \multicolumn{1}{c}{\multirow{6}*{\rotatebox[origin=c]{90}{Constraint}}} & No Mask & 0.542 & 0.568 & 0.580 & - \\
    &NE Del & 0.531 & 0.588 & 0.570 & - \\
    &Basic NER & 0.543 & 0.568 & 0.580 & - \\
    &WikiD & *\textbf{0.556} & 0.607 & 0.570 & - \\
    &WikiD+Del & 0.544 & \textbf{0.627} & \textbf{0.590} & - \\
    &WikiD+NER & 0.549 & 0.607 & 0.570 & - \\ \bottomrule
    \end{tabularx}
    }
    \caption{\label{out_domain}  Accuracy of out-of-domain data.
    The left-most column lists the training set, and each column with accuracy corresponds to each test set.
    We applied the statistical significance test by McNemar’s test~\cite{test} with Bonferroni correction to each method compared to No Mask. * indicates the significant difference over No Mask ($p < 0.05$).}
\end{table}


\section{Conclusion}
This study proposed a new bias concept, Diachronic Bias, caused by the difference in the creation period of various fake news datasets.
We firstly examined the deviation of phrase appearance in respective fake news detection datasets with different creation periods.
We then proposed masking methods using Wikidata to mitigate the influence of person names.
These masking methods achieved higher accuracy in out-of-domain datasets and showed to be made a model more robust.

In the future, more sophisticated approaches such as utilizing a knowledge graph to mitigate diachronic bias would be considered for fake news detection models.
In addition to diachronic bias, political bias and racial bias are likely to exist in fake news detection datasets; clarifying these biases in detail is an important next research direction.



\bibliography{anthology,custom}
\bibliographystyle{acl_natbib}

\clearpage

\appendix

\section{Overview of datasets}
\label{dataset}
The domains and the number of samples for each label in each dataset are listed in Table~\ref{tab:Dataset}.
We do not use the validation set in our experiments because the number of samples in Horne17 and Celebrity is small.

\begin{table}[h]
    \centering
    
    \caption{Overview of datasets}
    \scalebox{0.85}{
    \begin{tabular}{lllrr}\toprule
        Dataset & Domain & Year & Real & Fake\\ \midrule
        MultiFC & Multi & 2007--2015 &3803 &4058\\
        Horne17 & Political & 2016 & 128 & 123\\
        Celebrity & Entertainment & 2016--2017 & 250 & 250\\
        Constraint & COVID-19 & 2020 &5600 & 5100\\ \bottomrule
    \end{tabular}
    }
    \label{tab:Dataset}
\end{table}

\section{LMI in Horne17 and Celebrity}
\label{lmi}
Table~\ref{LMI_appendix} shows the top LMI-ranked bi-grams that are highly correlated with each label in Horne17 and Celebrity.
The result indicates that fake labels in Horne17 and Celebrity have a high correlation with the celebrity's name, such as ``brad pitt'' and ``kate middleton'' as well as  ``donald trump'' and ``hillary clinton.''
On the other hand, real labels have a high correlation with common phrases such as ``i had'' and ``would be.” 
These four datasets have a tendency to be highly correlated between person names and fake labels and between common phrases and real labels.


\begin{table}[t]
    \centering
    \scriptsize
    \scalebox{0.9}{
    \begin{tabular}{lrrlrr} \toprule
         \multicolumn{6}{c}{Horne17}\\ \midrule
         \multicolumn{3}{c}{Real} & \multicolumn{3}{c}{Fake}\\ \cmidrule(lr){1-3} \cmidrule(lr){4-6}
         \textbf{Bi-gram} & \textbf{LMI} & $p(l|w)$ & \textbf{Bi-gram} & \textbf{LMI} & $p(l|w)$\\ \cmidrule(lr){1-3} \cmidrule(lr){4-6}
         \textbf{trump has} & 112 & 0.82 & \textbf{donald trump} & 605 & 0.42\\
         national security & 106 & 0.88 & \textbf{hillary clinton} & 440 & 0.50\\
         would be & 104 & 0.72 & i think & 292 & 0.68\\
         people who & 92 & 0.89 & united states & 258 & 0.51\\
         transition team & 88 & 1.0 & have been & 230 & 0.41\\
         \textbf{mr. trump} & 80 & 0.94 & \textbf{bill clinton} & 208 & 0.70\\
         smug style & 77 & 1.0 & we are & 206 & 0.56\\
         \textbf{george w.} & 76 & 0.90 & \textbf{hillary clinton's} & 187 & 0.58\\
         republican party & 76 & 0.91 & \textbf{president obama} & 171 & 0.55\\
         new york & 70 & 0.77 & \textbf{ted cruz} & 149 & 0.80\\ \bottomrule
         \toprule
         \multicolumn{6}{c}{Celebrity}\\ \midrule
         \multicolumn{3}{c}{Real} & \multicolumn{3}{c}{Fake}\\ \cmidrule(lr){1-3} \cmidrule(lr){4-6}
         \textbf{Bi-gram} & \textbf{LMI} & $p(l|w)$ & \textbf{Bi-gram} & \textbf{LMI} & $p(l|w)$\\ \cmidrule(lr){1-3} \cmidrule(lr){4-6}
         i think & 233 & 0.90 & has been & 343 & 0.55\\
         i don\'t & 164 & 0.95 & do think & 214 & 0.80\\
         they were & 102 & 0.70 & an insider & 199 & 0.88\\
         i had & 100 & 0.94 & \textbf{brad pitt} & 163 & 0.63\\
         so i & 100 & 0.92 & insider told & 157 & 0.90\\
         but i & 87 & 0.79 & may have & 128 & 0.85\\
         we were & 87 & 0.89 & \textbf{kate middleton} & 124 & 0.88\\
         what i & 75 & 0.87 &  they are & 122 & 0.51\\
         i love & 70 & 0.92 & \textbf{the weeknd} & 119 & 0.62\\
         when i & 69 & 0.92 & \textbf{kanye west} & 113 & 0.56\\ \bottomrule
    \end{tabular}
    }
    \caption{Top 10 LMI ranked bi-grams in Horne17 and Celebrity datasets for real and fake labels with their $p(l | w)$.
    \textbf{LMI} are written as value multiplied by $10^{6}$.
    Person names are written in bold.}
    \label{LMI_appendix}
\end{table}

\section{Wikidata for masking methods}
\label{wikidata}
We utilized Wikidata corresponding to the creation time of the articles and posts in each dataset.
Specifically, we utilized Wikidata released on January 4, 2016, for MultiFC, January 15, 2018, for Horne17 and Celebrity, and December 28, 2020, for Constraint.

\section{Examples of outputs by each masking method}
\label{masking_ref}
Table~\ref{masking} lists outputs by the masking methods.

\begin{table*}[t]
    \centering
    \footnotesize
    \begin{tabularx}{\linewidth}{lX}
        \toprule 
        No Mask & 18 states including \textbf{US} \textbf{UK} and \textbf{Australia} request PM \fbox{\textbf{Modi}} to head a task force to stop coronavirus\\
        NE Del & 18 states including and request PM to head a task force to stop coronavirus\\
        Basic NER & 18 states including \textbf{LOC LOC} and \textbf{LOC} request PM \fbox{\textbf{PER}} to head a task force to stop coronavirus\\
        WikiD & 18 states including \textbf{US} \textbf{UK} and \textbf{Australia} request PM \fbox{\textbf{Q22337580}} to head a task force to stop coronavirus\\
        WikiD+Del & 18 states including and request PM \fbox{\textbf{Q22337580}} to head a task force to stop coronavirus\\
        WikiD+NER & 18 states including \textbf{LOC LOC} and \textbf{LOC} request PM \fbox{\textbf{Q22337580}} to head a task force to stop coronavirus\\
    \bottomrule
    \end{tabularx}
    \caption{\label{masking}
    Example fake news with the application of masking methods.
    \fbox{PER label} is represented by squares and other \textbf{NE label} is represented in bold. In Wikidata, Q22337580 indicates Chief Minister of Gujarat.}
\end{table*}

\begin{table*}[!t]
    \centering
    \footnotesize
    \begin{tabularx}{\linewidth}{lXl}
        \toprule 
        Dataset & Text & Label\\ \midrule
        Horne17 & \textbf{Trump} wins Electoral College vote as insurgency fizzles WASHINGTON - \textbf{Donald Trump} will - officially - become president next month. Trump surpassed the 270 electoral votes ...& Real\\ 
        Constraint & How many FROM covid 19? How many died because New York and New Jersey screwed the elderly?? Thats all \textbf{trumps} fault right? When \textbf{trump} shut down travelhe a racist \textbf{Trump} puts a team together to figure out the virusits not diverse enough & Fake\\
        Constraint & \textbf{AG Barr} Suggests an End to the Coronavirus Lockdown URL & Fake\\
    \bottomrule
    \end{tabularx}
    \caption{\label{tweet_table}
    Some examples, which WikiD trained on MultiFC accurately classifies while No Mask trained on MultiFC makes the wrong classification. These examples include politicians such as ``Donald Trump'' and ``William Barr''.}
\end{table*}

\begin{table}[h]
    \centering
    \small
    \begin{tabular}{lrrrr} \toprule
         & MultiFC & Horne17 & Celebrity & Constraint\\ \midrule
         MultiFC & - & 37.4\% & 29.4\% & 30.1\%\\
         Horne17 & 53.6\% & - & 36.4\% & 39.9\%\\
         Celebrity & 38.9\% & 33.6\% & - & 33.6\%\\
         Constraint & 35.3\% & 34.1\% & 28.6\% & -\\
         \bottomrule
    \end{tabular}
    \caption{Coverage rate of Wikidata label between each dataset. These percentage values indicate how much of Wikidata labels in the data listed in the left-most column is contained in other datasets.}
    \label{percentile}
\end{table}

\section{Relationship between Wikidata labels in each dataset}
\label{wiki_ref}
Table~\ref{percentile} presents the coverage rate of Wikidata label between each dataset.
Each percentage value represents the number of Wikidata labels in the dataset in the left column covered by Wikidata labels in other datasets.
The MultiFC and Horne17 datasets have a high coverage rate because they contain articles and posts on the same political topics.
On the other hand, the coverage of the Celebrity dataset is lower than that of the other datasets owing to the entertainment domain.
Table~\ref{top_words} presents the top-3 appearance ranked Wikidata label in each dataset.

\begin{table}[h]
    \small
    \scalebox{0.8}{
    \begin{tabular}{crrrr} \toprule
         Rank & MultiFC & Horne17 & Celebrity & Constraint\\ \midrule
         \multirow{2}*{1} & President of & President of &  \multirow{2}*{actor} & President of\\
         & the U.S. & the U.S. & & the U.S.\\ \midrule
         \multirow{2}*{2} & U.S. & Attorney General & \multirow{2}*{singer} & \multirow{2}*{CEO}\\
         & representative & of Arkansas & & \\ \midrule
         \multirow{2}*{3} & Secretary & Secretary & television & Mayor\\
         & of State & of State & actor & of London\\
         \bottomrule
    \end{tabular}
    }
    \caption{Top-3 ranked appearances in Wikidata label of each dataset}
    \label{top_words}
\end{table}

\section{Implementation Details}

\paragraph{Hyperparameters}
Hyperparameters for training our model are as below: learning rate is $1.0 \times 10^{-5}$, batch size is $16$ and sentence length is $512$.
We selected the initial value in huggingface library (one of the python libraries) as hyperparameters, based on the empirical rule that fine-tuning works well in our dataset.

\paragraph{Training Efficiency}
Since our model has the same architecture as BERT~\cite{DBLP:journals/corr/abs-1810-04805} except for new tokens that are added for new labels, it holds approximately 110M parameters.
We use a Quadro RTX 8000 GPU to train our model.
Training our model takes 3 epochs (about 1 hour) for MultiFC and Constraint datasets, and 8 epochs (about 20 minutes) for Horne17 and Celebrity datasets.

\section{Experiments on time-based splitting of MultiFC}
\label{time_ref}
Not only in-domain and out-of-domain experimental settings, but we also conduct an experiment based on time-based splitting of MultiFC into train and test sets.
The time-based experimental setting is similar to the intent of out-of-domain experimental settings (refer to Sec.4.3.2).
It intends to verify the effect of each masking method in the same dataset for trying to remove more of the effects of domain shift.
The experimental setting can not be applied to other datasets because only MultiFC has the published time of each sample.
We set samples published in 2012--2015 as train set (3508 samples) and samples published in 2016--2018 as test set (2389 samples).

Table~\ref{time_exp} presents the accuracies of each masking method against MultiFC test set (2016-2018).
Same as out-of-domain experiments, almost masking methods achieved higher accuracy than No Mask.
This show that the masking method using Wikidata could mitigate the diachronic bias and build robust models even for time-based splitting of a dataset.

\begin{table}[t]
    \centering
    \footnotesize
    \begin{tabular}{lr} \toprule
    \multicolumn{1}{c}{Method} & Test set: MultiFC (2016-2018)\\ \midrule
    No Mask & 0.639\\
    NE Del & 0.659\\
    Basic NER & 0.639\\
    WikiD & *0.664\\
    WikiD+Del & *\textbf{0.671}\\
    WikiD+NER & 0.644\\
    \bottomrule
    \end{tabular}
    \caption{\label{time_exp}
    Accuracy of each masking method against MultiFC test set (2016-2018). Bold indicates the highest accuracy. }
\end{table}

\section{Some examples of dataset}
\label{example_tweet}
In Table~\ref{tweet_table}, we show some examples, which WikiD trained on MultiFC accurately classifies while No Mask trained on MultiFC makes the wrong classification.

\end{document}